\documentclass[10pt,conference]{IEEEtran}

\IEEEoverridecommandlockouts

\usepackage{cite}
\usepackage{amsmath,amssymb,amsfonts}
\usepackage{graphicx}
\usepackage{textcomp}
\usepackage{xcolor}
\usepackage{comment}
\usepackage{booktabs}
\usepackage{multirow}
\usepackage{tabularx}
\usepackage{mathtools}
\usepackage{caption}
\usepackage[normalem]{ulem}
\usepackage{soul}
\usepackage{algorithm}
\usepackage{algpseudocode}

\def\BibTeX{{\rm B\kern-.05em{\sc i\kern-.025em b}\kern-.08em
    T\kern-.1667em\lower.7ex\hbox{E}\kern-.125emX}}
\begin{document}
\bstctlcite{IEEEexample:BSTcontrol}
\title{Wireless Edge-Assisted Adaptive Latent Reasoning for Vision-Language-Action Models in Physical AI}

\author{\IEEEauthorblockN{Fatih Temiz and Melike Erol-Kantarci, \textit{Fellow, IEEE}}
\IEEEauthorblockA{\textit{School of Electrical Engineering and Computer Science, University of Ottawa, Ottawa, Canada} \\
Emails: \{ftemi033, melike.erolkantarci\}@uottawa.ca}}

\maketitle

\begin{abstract}
Vision-language-action (VLA) models enable embodied artificial intelligence (AI) by translating visual observations and language instructions into robotic actions. Emerging use cases in physical AI are expected to drive the need for efficient inference and communication, as the computational demands of the VLA models pose challenges for real-time control with limited onboard resources. Test-time scaling through latent reasoning allows these models to refine hidden representations without generating explicit chain-of-thought (CoT) tokens, but introduces a trade-off between task success, inference latency, and energy consumption. In this paper, we address the efficient collaborative inference under a realistic communication scenario. We propose a wireless edge-assisted framework for adaptive latent reasoning (WEALR) that augments onboard computation with edge computing. A decision-making agent jointly selects the execution location and latent-reasoning convergence threshold. We formulate this joint adaptation as a constrained Markov decision process (CMDP). 
A Lagrangian-constrained deep reinforcement learning (CDRL) approach, namely trust region policy optimization Lagrangian (TRPO-Lag), learns to balance task performance and resource consumption under varying wireless conditions and task requirements. We evaluate the framework using LIBERO manipulation tasks, profiled VLA inference costs, and ray-traced wireless channels. Experimental results demonstrate that TRPO-Lag achieves up to 23\% higher feasibility-weighted task success rate compared to benchmark CDRL methods. 
\end{abstract}
\vspace{-10pt}
\section{INTRODUCTION}
Physical artificial intelligence (AI) brings perception, reasoning and action into embodied systems that interact with the physical world 
\cite{wang2026phyairealtimephysicalai}. Vision-language-action (VLA) models, such as OpenVLA \cite{openvla}, support this capability by translating visual observations and language instructions into robotic actions. Advances in these models create opportunities for generalizable physical AI, but their computational demands challenge deployment on resource-constrained robots. Task performance must therefore be balanced against onboard computation, energy consumption, and latency requirements \cite{jiang2026fastirunvla,zhou2026characterizing}. In particular, timely robotic control requires communication and inference to jointly satisfy a latency service-level objective (SLO). We refer to robots capable of visual perception, instruction understanding, planning, reasoning, and action execution as physical AI devices (PADs) in this paper. 

Edge computing augments limited onboard computation by enabling PADs to offload inference to nearby edge servers, but its benefits depend on wireless propagation conditions \cite{acm_survey_ai_reasoning}. 
This deployment perspective aligns with the vision of using radio access network (RAN) infrastructure to support AI applications at the network edge, such as mobile switching offices (MSO) \cite{11141658}.
However, faster edge computation comes at the cost of wireless transmission delay. In indoor environments such as homes and warehouses, blockage and multipath propagation create heterogeneous line-of-sight (LOS) and non-line-of-sight (NLOS) conditions. Changes in PAD position affect achievable uplink (UL) rates and can offset the computational advantage of edge execution. Effective offloading must therefore account for wireless conditions alongside computation and control-latency requirements. Evaluating this tradeoff requires environment-specific channel models that capture blockage, multipath propagation, and spatial variations in link quality, which ray tracing can provide \cite{alkhateeb2019deepmimo, farzanullah2025wirelessmultimodalfoundationmodel}.

Reliable physical interaction requires VLA models to interpret diverse visual observations and language instructions and generate appropriate actions across tasks of varying difficulty. Supporting these capabilities motivates increasing model capacity through parameter scaling or allocating additional computation during inference.
Sparse mixture-of-experts (MoE) architectures enable parameter scaling while limiting computation per input by activating only a subset of model parameters \cite{temiz2026wisdomwirelesssparsedecision}. However, their memory and computational requirements can still challenge deployment on resource-constrained PADs. Test-time scaling offers a complementary approach that adjusts computational effort within a fixed model, making it particularly attractive for wireless edge deployment, where available resources and latency requirements limit inference costs \cite{acm_survey_ai_reasoning}.

Unlike explicit reasoning methods such as chain-of-thought (CoT), which generate intermediate text through sequential decoding, latent reasoning refines hidden representations without producing intermediate text tokens, potentially reducing the decoding overhead and latency associated with textual reasoning. The recurrent-depth architecture in~\cite{geiping2025scaling} enables large language models (LLMs), which serve as backbones of VLAs, to vary computational depth through shared parameters, offering adjustable inference effort within a fixed model footprint. For robotic inference, recurrent-depth VLA (RD-VLA) applies a related principle through a weight-tied recurrent action head and a convergence-based stopping criterion to adapt the number of reasoning iterations across observations \cite{tur2026rdvla}. For example, grasping a small object may
require more reasoning than moving through an unobstructed
region. A single convergence threshold can therefore yield different reasoning depths across stages of a task and different tasks. Additional reasoning may improve action quality but also increases inference latency and resource consumption, motivating adaptive control of reasoning effort.
These considerations motivate jointly selecting VLA execution location and reasoning effort under wireless, energy, and computation constraints. 

Related studies address complementary aspects of collaborative inference. The device-cloud framework in~\cite{devicecloudmultiturnllm} uses resource-constrained reinforcement learning (RL) to select LLM inference locations and multimodal inputs. Also, the EcoVLA approach proposed in \cite{ecovla} selects energy-efficient VLA
co-inference schemes under real-time constraints,
adapting to network fluctuations and edge workload
variations. The authors of~\cite{liu2026mores} propose MORES, a \textit{reasoning-as-a-service} framework that distributes recurrent hidden-state updates between devices and servers, jointly adapting reasoning computation and latent-feature pruning under heterogeneous wireless conditions. 
While conversational LLM services typically prioritize response quality, VLAs must deliver actions within deadlines imposed by ongoing physical control \cite{jiang2026fastirunvla}. 

Meeting these deadlines requires balancing task success against communication delay and resource consumption. 
Cumulative budgets couple decisions across requests, and shared edge resources couple decisions across PADs, motivating a constrained Markov decision process (CMDP) formulation. Constrained deep reinforcement learning (CDRL) addresses such problems by separating performance objectives from explicit operating limits~\cite{ray2019benchmarking}. This motivates the use of CDRL methods, including Lagrangian variants of proximal policy optimization (PPO) and trust region policy optimization (TRPO), as well as constrained policy optimization (CPO), for resource management in wireless networks. In particular, PPO-Lag has been adopted for joint unmanned aerial vehicle (UAV) altitude control and Internet of Things (IoT) channel access~\cite{9177252}.

In this paper, we propose Wireless Edge-Assisted Adaptive Latent Reasoning (WEALR), a framework that jointly adapts VLA execution location and reasoning effort. For each robotic task request, an agent selects local or edge execution and a reasoning convergence threshold. This threshold governs termination of the latent reasoning loop while allowing the realized reasoning depth to vary across observations within the task.
We use TRPO-Lag, which learns the policy and constraint multipliers, connecting adaptive latent reasoning with wireless offloading and cumulative resource management.
The key contributions are as follows:
\begin{enumerate}
    \item We develop WEALR to jointly select VLA execution location and reasoning threshold according to task requirements, wireless conditions, and available resources.
    \item We formulate a CMDP with per-PAD energy, shared edge-reasoning, and SLO constraints, and apply TRPO-Lag to optimize robotic task success.
    \item We evaluate task success, feasibility, and policy behavior using LIBERO manipulation tasks \cite{liu2023libero}, profiled VLA inference costs, and ray-traced channels under varying wireless conditions and resource budgets.
\end{enumerate}

\begin{figure*}[h!]
    \centering
    \includegraphics[width=0.95\textwidth]{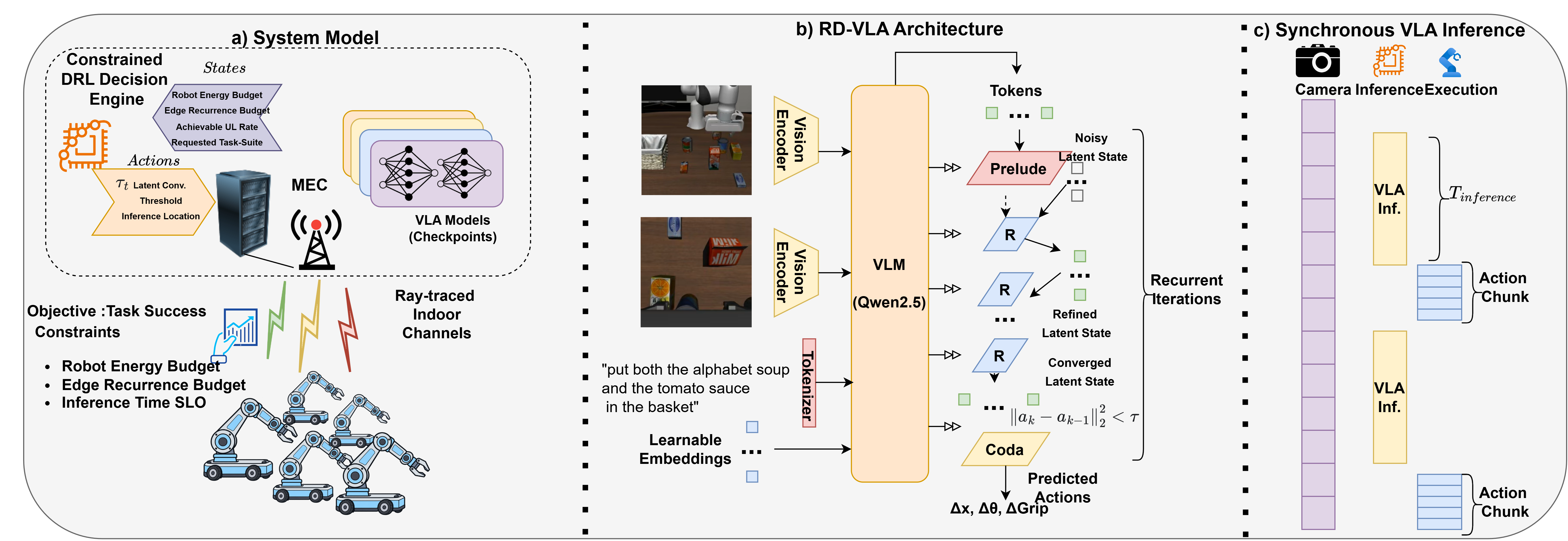}
    \caption{Overview of the proposed WEALR framework.}
    \label{fig:WEALR_framework}
\end{figure*}

\begin{figure}[h!]
    \centering
    \includegraphics[width=0.85\columnwidth]{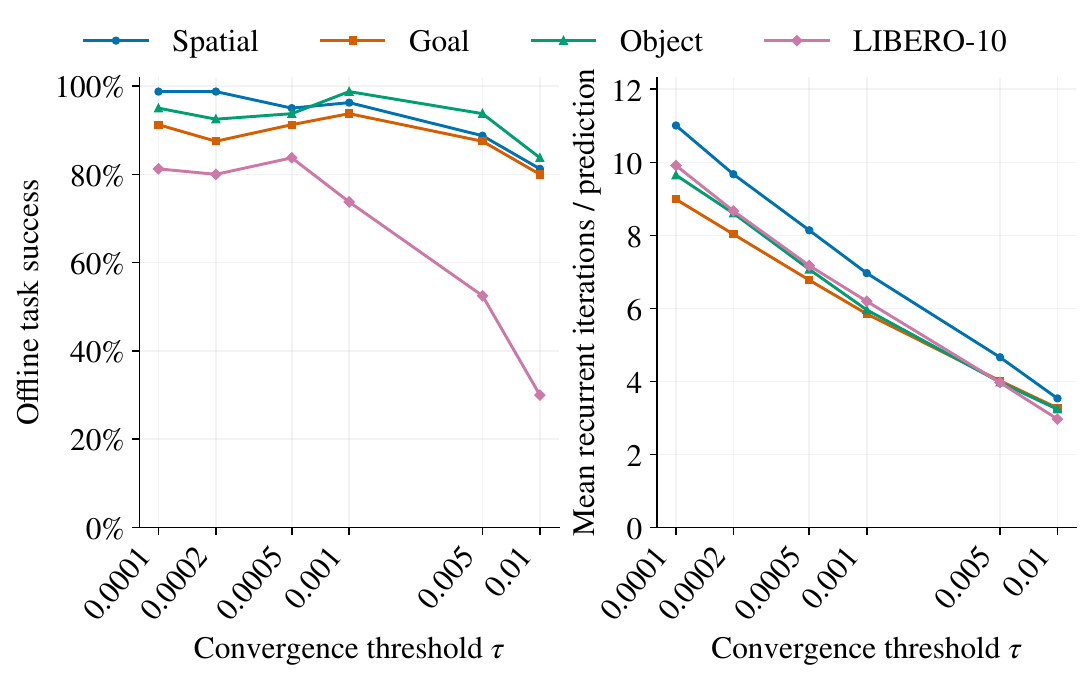}
    \caption{VLA task performance versus inference cost}
    \label{fig:offlinedatasetcharacteristics}
    \vspace{-0.5cm}
\end{figure}

\section{SYSTEM MODEL AND PROBLEM FORMULATION}
\label{sec:system_model}

We consider an edge-computing system comprising $N$ PADs
and an edge server each equipped with AI acceleration hardware co-located
with an indoor base station (BS), as shown in Fig.~\ref{fig:WEALR_framework} (a).
Both PADs and the server host pretrained VLA models
with preloaded checkpoints; cold-start latency is neglected \cite{ecovla}.
Each episode comprises $K$ requests, with request $k$
originating from PAD $i_k$ and belonging to task suite
$q_k$. The controller selects an execution location and
reasoning threshold for each request, fixed throughout the task.
Requests are served sequentially, with only one active
UL transmission at a time~\cite{liu2026mores}. We use synchronous VLA inference: each cycle begins after the previous action chunk is executed, avoiding potentially
stale observations in asynchronous execution~\cite{jiang2026fastirunvla}.
\subsection{System Model}

\subsubsection{Latent Reasoning for VLAs}
\label{subsubsec:latenreasoning}

The VLA processes visual observations and language
instructions to generate action chunks, each comprising
a sequence of actions (e.g., five). During each inference
cycle, recurrent updates refine the hidden representations
until the predicted action chunk stabilizes, as illustrated
in Fig.~\ref{fig:WEALR_framework}(b) and (c). Following
RD-VLA, convergence is assessed using the squared
Euclidean distance between consecutive predicted action
chunks as a proxy for the Kullback-Leibler (KL) divergence between their
corresponding action distributions:
\begin{equation}
\Delta_{k,m}^{(r)}
=\|\mathbf{a}_{k,m}^{(r)}
-\mathbf{a}_{k,m}^{(r-1)}\|_2^2,
\end{equation}
\begin{equation}
d_{k,m}(\tau_k)
=\min\left(
\{r:\Delta_{k,m}^{(r)}<\tau_k\}
\cup\{d_{\max}\}
\right).
\end{equation}
where $\mathbf{a}_{k,m}^{(r)}$ denotes the predicted action
chunk at recurrent update $r$, $m$ indexes inference cycles,
and $d_{\max}$ is the maximum reasoning depth. Thus, a
single threshold governs each task, while the realized
depth varies across observations.
Unlike fixed-depth reasoning~\cite{liu2026mores}, a task-level
threshold adapts depth across observations, potentially
reducing recurrent updates while maintaining task
success ~\cite{tur2026rdvla}.

An offline dataset profiles task success, inference latency,
energy, and recurrent-update counts across task suites and
candidate thresholds, as summarized in
Fig.~\ref{fig:offlinedatasetcharacteristics}. 
Increasing the convergence threshold reduces the mean recurrent iterations per inference across all suites, but generally lowers task success, with the largest degradation observed in LIBERO-10, which involves long-horizon tasks \cite{liu2023libero}.
These tradeoffs motivate adaptive threshold selection
and provide the profiles used by the resource-management
environment.

\subsubsection{Communication Model}

We consider an orthogonal frequency division multiplexing
(OFDM) UL channel in which single-antenna PADs communicate
with a BS equipped with $M$ antennas. Indoor channels are
generated using DeepMIMO by mapping each PAD's position
to the nearest channel location, capturing geometry-dependent
blockage and LOS/NLOS multipath
propagation~\cite{alkhateeb2019deepmimo}.
PAD mobility follows the random waypoint model.
For request $k$, the channel of PAD $i_k$ on subcarrier
$n$ is $\mathbf{h}_{i_k,k}[n]\in\mathbb{C}^{M}$.
To represent bandwidth availability under exogenous
traffic at the BS, $B_k$ is sampled uniformly from a
predefined set $\mathcal{B}$ for each request and mapped to a central contiguous
subcarrier set $\mathcal{L}_k$.
The total transmit power $P_{\mathrm{tx}}$ is allocated
equally across the active subcarriers.
With maximum-ratio combining, the UL signal-to-noise ratio (SNR) and achievable
rate are:
\begin{align}
\gamma_{i_k,k}[n]
&=\frac{P_{\mathrm{tx}}
\|\mathbf{h}_{i_k,k}[n]\|_2^2}
{L_kN_0F_{\mathrm{NF}}\Delta f}, \\
R_{i_k,k}^{\mathrm{UL}}
&=\Delta f\sum_{n\in\mathcal{L}_k}
\log_2\!\left(1+\gamma_{i_k,k}[n]\right),
\end{align}
where $\mathcal{L}_k$ is the active subcarrier set,
$L_k=|\mathcal{L}_k|$, $\Delta f$ is the subcarrier
spacing, $N_0$ is the noise spectral density, and
$F_{\mathrm{NF}}$ is the linear receiver noise figure.
Only the requesting PAD transmits, excluding
interference among the modeled PADs.

Each inference cycle uploads two uncompressed
$224\times224$ RGB images ($D_{\mathrm{UL}}\approx2.41$~Mbit),
neglecting prompt and proprioceptive payloads.
The upload latency and transmit energy for inference
cycle $m$ of request $k$ are:
\begin{equation}
T_{k,m}^{\mathrm{UL}}
=\frac{D_{\mathrm{UL}}}{R_{i_k,k}^{\mathrm{UL}}},
\qquad
E_{k,m}^{\mathrm{UL}}
=P_{\mathrm{tx}}T_{k,m}^{\mathrm{UL}}.
\end{equation}
Downlink transmission delay and receive energy are
neglected due to the small returned action-chunk payload (e.g., 140 bytes).
We assume that the PAD remains stationary during each task execution and the UL rate remains fixed throughout each task.

\subsubsection{Computation Model}
Let $\ell_k=0$ and $\ell_k=1$ denote local and edge
execution, respectively. The inference latencies
$T_{k,m}^{\mathrm{loc}}(\tau_k)$ and
$T_{k,m}^{\mathrm{edge}}(\tau_k)$ account for the
respective hardware capabilities and selected convergence
threshold. The total chunk latency is
\begin{equation}
T_{k,m}=(1-\ell_k)T_{k,m}^{\mathrm{loc}}(\tau_k)
+\ell_k\left[T_{k,m}^{\mathrm{UL}}
+T_{k,m}^{\mathrm{edge}}(\tau_k)\right].
\end{equation}
To account for the server's finite computation capacity
shared among PADs, edge workload is measured in
reasoning iterations~\cite{liu2026mores}.
PAD energy and edge reasoning consumption for task $k$
are
\begin{align}
e_k&=\sum_{m=1}^{M_k}\left[
(1-\ell_k)E_{k,m}^{\mathrm{loc}}(\tau_k)
+\ell_kE_{k,m}^{\mathrm{UL}}\right], \\
g_k&=\ell_k\sum_{m=1}^{M_k}d_{k,m}(\tau_k),
\end{align}
where $M_k$ is the number of inference chunks,
$E_{k,m}^{\mathrm{loc}}(\tau_k)$ and
$E_{k,m}^{\mathrm{UL}}$ denote PAD inference and
UL transmission energy, respectively, and
$d_{k,m}(\tau_k)$ is the reasoning iteration count. Locomotion energy is excluded, since PAD motion is independent of the inference execution location.
Cumulative energy for PAD $i$ and shared edge
consumption are
$E_i=\sum_k\mathbf{1}\{i_k=i\}e_k$ and
$G_{\mathrm{edge}}=\sum_k g_k$.

A task violates the SLO if more than a fraction $\eta$
of its chunks exceed the deadline $T_{\mathrm{SLO}}$:
\begin{equation}
z_k=\mathbf{1}\left\{
\frac{1}{M_k}\sum_{m=1}^{M_k}
\mathbf{1}\{T_{k,m}>T_{\mathrm{SLO}}\}>\eta
\right\}.
\end{equation}

\subsection{Problem Formulation and CMDP Framework}

\subsubsection{Problem Statement}
We maximize expected task success subject to PAD energy,
edge reasoning, and SLO constraints:
\begin{subequations}\label{eq:cmdp}
\begin{equation}
\max_{\pi_\theta}\quad
\mathbb{E}_{\pi_\theta}
\left[\sum_{k=1}^{K}y_k\right]
\tag{\theparentequation}\label{eq:cmdp_objective}
\end{equation}
\begin{equation}\label{eq:cmdp_energy}
\mathrm{s.t.}\quad
\mathbb{E}_{\pi_\theta}
\left[\max_i\frac{E_i}{E_{\max}}\right]\leq1,
\end{equation}
\begin{equation}\label{eq:cmdp_edge}
\mathbb{E}_{\pi_\theta}
\left[\frac{G_{\mathrm{edge}}}{G_{\max}}\right]\leq1,
\end{equation}
\begin{equation}\label{eq:cmdp_slo}
\mathbb{E}_{\pi_\theta}
\left[\frac{1}{K}\sum_{k=1}^{K}z_k\right]\leq\epsilon.
\end{equation}
\end{subequations}
where $y_k=1$ if the PAD successfully completes the
intended task. Here, $E_i$ is PAD $i$'s total
energy consumption, $G_{\mathrm{edge}}$ is the total edge
recurrence count, and $E_{\max}$ and $G_{\max}$ are their
respective budgets. 
Constraints bound expected
episode costs; episode feasibility is assessed using
realized costs.

\subsubsection{CMDP Formulation}
The underlying system is modeled as a finite-horizon CMDP
with $K$ decision steps, one per task request. The policy
$\pi_\theta(a_k\mid\mathbf{o}_k)$ specifies action
probabilities given a nine-dimensional observation.
Since the individual remaining energies of non-requesting
PADs are omitted, this observation provides an
approximate state representation.

\paragraph{Observation Space}
Before serving request $k$, the agent observes
\begin{equation}
\mathbf{o}_k=\left[
\mathbf{q}_k,\,
\frac{E_{i_k,k}^{\mathrm{rem}}}{E_{\max}},\,
\frac{G_k^{\mathrm{rem}}}{G_{\max}},\,
R_{i_k,k}^{\mathrm{UL}},\,
\frac{k}{K},\,
\max_{i\in\mathcal{N}}\frac{E_{i,k}^{\mathrm{used}}}{E_{\max}}
\right],
\end{equation}
where $\mathbf{q}_k$ is a four-dimensional one-hot task-suite encoding (Spatial, Object, Goal, or LIBERO-10). Here, $E_{i_k,k}^{\mathrm{rem}} = E_{\max} - E_{i_k,k}^{\mathrm{used}}$ denotes the remaining energy of requesting PAD $i_k$, while the last term represents the maximum consumed-energy ratio across all PADs. In addition, $G_k^{\mathrm{rem}}=G_{\max}-\sum_{t=1}^{k-1}g_t$ is the remaining edge reasoning budget, where $g_t$ denotes the edge recurrence count for request $t$. 

\paragraph{Action Space}
Each action selects an execution location
and a convergence threshold, both fixed throughout the task:
\begin{equation}
\begin{aligned}
a_k&=(l_k,\tau_k),\qquad l_k\in\{0,1\},\\
\tau_k&\in\{10^{-4},2{\times}10^{-4},5{\times}10^{-4},
10^{-3},5{\times}10^{-3},10^{-2}\},
\end{aligned}
\end{equation}
yielding 12 possible actions.

\paragraph{Reward and Costs}
Upon task completion, the reward is $r_k=y_k$ and
the costs are
\begin{align}
c_k^{\mathrm{E}}
&=\frac{\max_i E_{i,k+1}^{\mathrm{used}}
-\max_i E_{i,k}^{\mathrm{used}}}{E_{\max}},\\
c_k^{\mathrm{G}}&=\frac{g_k}{G_{\max}},\qquad
c_k^{\mathrm{S}}=\frac{z_k}{K}.
\end{align}
These costs measure PAD energy, edge reasoning, and
SLO violations, respectively. With
$E_{i,1}^{\mathrm{used}}=0$ and
$E_i=E_{i,K+1}^{\mathrm{used}}$, their episode sums
recover the quantities in
\eqref{eq:cmdp_energy}--\eqref{eq:cmdp_slo}.

\section{CDRL-Based Adaptive Latent Reasoning}

\subsection{TRPO-Lagrangian}

We adopt TRPO-Lag to solve \eqref{eq:cmdp},
combining trust-region policy updates with learned
constraint penalties~\cite{pmlrschulman15,ray2019benchmarking}.
Given observation $\mathbf{o}_k$, the actor samples
$a_k=(l_k,\tau_k)$ from the 12 available actions.
The selected configuration is maintained throughout
the task, after which success and resource costs
are recorded.

Let $\mathcal{J}=\{\mathrm{E},\mathrm{G},\mathrm{S}\}$
index the constraints, with cumulative costs
$C_j=\sum_{k=1}^{K}c_k^j$ and limits
$(d_{\mathrm{E}},d_{\mathrm{G}},d_{\mathrm{S}})
=(1,1,\epsilon)$.
The policy maximizes the Lagrangian objective:
\begin{equation}
\mathcal{L}(\theta,\boldsymbol{\lambda})
=
\mathbb{E}_{\pi_\theta}\!\left[\sum_{k=1}^{K}r_k\right]
-\sum_{j\in\mathcal{J}}\lambda_j
\frac{\mathbb{E}_{\pi_\theta}[C_j]-d_j}{s_j},
\end{equation}
where $\lambda_j\geq0$ is an adaptive multiplier and
$s_j>0$ scales the corresponding cost.
Unlike manual reward shaping with fixed weights, this formulation dynamically adapts the penalties to satisfy the constraints directly \cite{9177252}.

To optimize \eqref{eq:cmdp}, we employ an alternating primal-dual scheme. 
In the primal step, a reward critic and three cost critics estimate future success and cumulative costs from $\mathbf{o}_k$.
Using generalized advantage estimation (GAE), we form
\begin{equation}
\widehat A_k^{\mathrm{Lag}}
=
\widehat A_k^{\mathrm{R}}
-\sum_{j\in\mathcal{J}}
\frac{\lambda_j}{s_j}\widehat A_k^j.
\end{equation}
This advantage is normalized over the rollout, yielding
$\widetilde A_k^{\mathrm{Lag}}$.
With the multipliers fixed, the actor update solves
the sampled trust-region problem
\begin{align}
\max_\theta\quad&
\widehat{\mathbb{E}}\!\left[
\frac{\pi_\theta(a_k\mid\mathbf{o}_k)}
{\pi_{\theta_{\mathrm{old}}}(a_k\mid\mathbf{o}_k)}
\widetilde A_k^{\mathrm{Lag}}\right]
\label{eq:trpo_update}\\
\mathrm{s.t.}\quad&
\widehat{\mathbb{E}}\!\left[
D_{\mathrm{KL}}\!\left(
\pi_{\theta_{\mathrm{old}}}(\cdot\mid\mathbf{o}_k)
\,\Vert\,\pi_\theta(\cdot\mid\mathbf{o}_k)
\right)\right]\leq\delta_{\mathrm{KL}}.
\nonumber
\end{align}
Here, $\widehat{\mathbb{E}}$ denotes the rollout average.
The KL bound limits changes in the probabilities of
offloading and threshold choices between updates.
We approximately solve \eqref{eq:trpo_update} using
damped conjugate gradients and backtracking, accepting
only steps that improve the surrogate objective and
satisfy the KL bound. The critics are fitted to their
respective GAE return targets.
\begin{figure*}[h!]
    \centering
    \includegraphics[width=1.9\columnwidth]{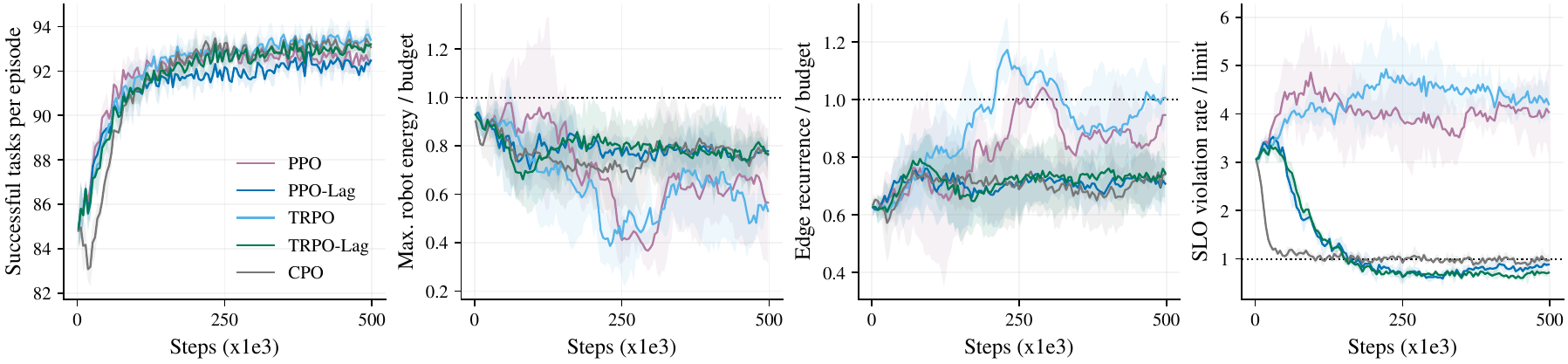}
    \caption{Training Performance and Normalized Constraint Costs}
    \label{fig:trainingperformance}
    \vspace{-0.5cm}
\end{figure*}
In the dual step, after each window of $W$ completed episodes, the multipliers are updated using a projected multiplier update:
\begin{equation}
\lambda_j\leftarrow
\left[\lambda_j+\alpha_\lambda
\frac{\widehat C_j-d_j}{s_j}\right]_+,
\qquad j\in\mathcal{J},
\label{eq:dual_update}
\end{equation}
where $\widehat C_j$ is the mean episode cost,
$s_j=d_j$ is the constraint-specific scaling factor,
$\alpha_\lambda$ is the dual learning rate, and
$[x]_+=\max(0,x)$.
Thus, exceeding a resource or SLO limit increases its
penalty, while operating below the limit reduces it.

\section{SIMULATION SCENARIOS AND RESULTS}

\subsection{Experimental Settings and Baselines}
\subsubsection{Experimental Settings}
We consider $N=5$ PADs served by an indoor BS with a co-located
edge server, with three PAD locations sampled from LOS
regions and two from NLOS regions. Each episode comprises
$K=100$ task requests. Methods are trained across three random seeds and evaluated deterministically over 100 held-out episodes. Table~\ref{tab:simulation_trpo_parameters} lists the
simulation parameters and TRPO-Lag hyperparameters. The offline dataset described in
Section~\ref{subsubsec:latenreasoning} comprises 2,400
RD-VLA rollouts across four LIBERO task suites, ten tasks
per suite, six convergence thresholds, and ten rollouts
per task--threshold configuration. Before training, rollouts are partitioned into fixed training (80\%) and test (20\%) subsets. Inference latency and
GPU power are profiled on an RTX~4090, with power sampled
through NVML and energy computed by trapezoidal integration.
To emulate Jetson Thor execution, profiled latency and energy are scaled by empirical factors $s_T$ and $s_E$, respectively
~\cite{zhou2026characterizing,jiang2026fastirunvla,11206288}.

We evaluate WEALR using an offline, profile-driven simulator.
At each decision step, a task request is sampled, and the
controller selects its execution location and reasoning threshold
based on observation $o_k$. A matching RD-VLA rollout is then sampled from
the corresponding dataset split. Its recorded task outcome
and action-chunk-level statistics are combined with hardware
cost scaling and the simulated wireless link to determine
task success, end-to-end latency, PAD energy consumption,
and edge reasoning cost.

\begin{table}[h]
\centering
\scriptsize
\setlength{\tabcolsep}{3pt}
\renewcommand{\arraystretch}{1.05}
\caption{TRPO-Lag and Simulation Parameters.}
\label{tab:simulation_trpo_parameters}
\begin{tabular}{@{}lclc@{}}
\toprule
\textbf{Simulation} & \textbf{Value}
& \textbf{TRPO-Lag} & \textbf{Value} \\
\midrule
DeepMIMO scenario & \texttt{I3\_2p4}
& Hidden layers & $[64,64]$ \\

$f_c$ & $2.4$ GHz
& Training steps & $5{\times}10^5$ \\

$M$ & $32$
& Rollout length & $1024$ \\

$\mathcal{B}$ & $\{1,2,5\}$ MHz
& Critic batch size & $64$ \\

$P_{\mathrm{tx}}$ & $20$ dBm
& Critic epochs & $10$ \\

$F_{\mathrm{NF}}$ & $7$ dB
& $\alpha_V$ & $3{\times}10^{-4}$ \\

$v$ (PAD Velocity)& $[0.5,1.5]$ m/s
& $\gamma$ & $1$ \\

$\epsilon$, $\eta$, $T_{SLO}$  & $0.05$, $0.1$, $200ms$
& $\lambda_{\mathrm{GAE}}$ & $0.95$ \\

$s_T$ & $1.69$
& $\delta_{\mathrm{KL}}$ & $0.01$ \\

$s_E$ & $0.535$
& CG iterations & $10$ \\

$E_{\max}$ & $2.27$ kJ
& $\alpha_\lambda$ & $0.05$ \\

$G_{\max}$ & $1.94{\times}10^4$ 
& $W$ & $40$ \\
\bottomrule
\end{tabular}
\end{table}
\subsubsection{Benchmark Methods}
We compare TRPO-Lag with:
\textbf{Task-aware heuristic (TAH)}, which selects a fixed
action per task suite using training-data estimates;
\textbf{Task and rate-aware heuristic (TRAH)}, which
additionally considers the current achievable UL rate;
\textbf{Random}, which uniformly samples the twelve actions;
\textbf{PPO and TRPO}, which maximize task success without
constraints~\cite{ray2019benchmarking,pmlrschulman15};
\textbf{CPO}, which incorporates all three constraints
into trust-region policy updates~\cite{ray2019benchmarking};
and \textbf{PPO-Lag}, which handles the same constraints
using adaptive Lagrange
multipliers~\cite{9177252,ray2019benchmarking}.

\subsection{Results}

In terms of CDRL performance, Fig. \ref{fig:trainingperformance} compares learning performance and normalized CMDP constraint costs. All methods improve task success during training, reaching approximately 92-94 successful tasks per episode, equivalent to the average episode reward. However, maximizing success alone does not ensure feasible operation. PPO and TRPO maintain SLO violation rates approximately four times the prescribed limit, while their edge reasoning costs approach or exceed the budget during training. In contrast, TRPO-Lag and PPO-Lag reduce SLO violations below the limit while maintaining comparable task success. TRPO-Lag stabilizes at approximately 0.8 of the maximum PAD energy budget and 0.7 of the edge reasoning budget. CPO also controls resource consumption, but its SLO violation rate remains close to the constraint boundary.
\begin{figure}[h!]
    \centering
    \includegraphics[width=1\columnwidth]{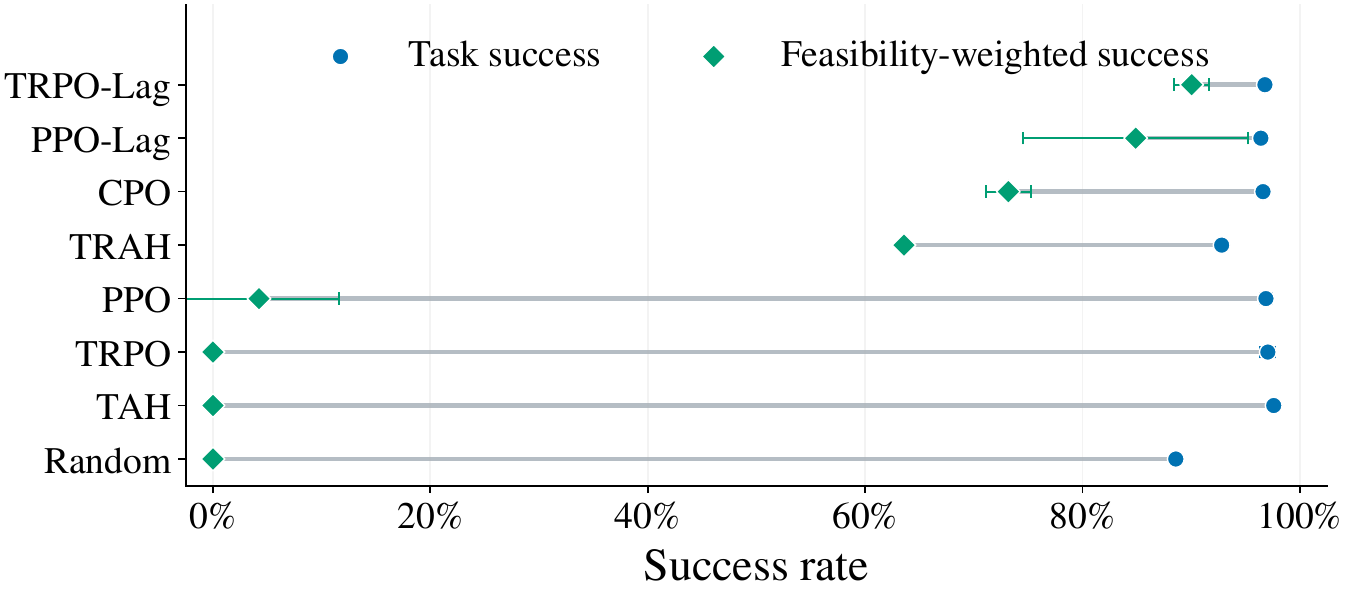}
    \caption{Task Success and Constraint-Aware Performance}
    \label{fig:ev}
    \vspace{-5pt}
\end{figure}
Fig. \ref{fig:ev} evaluates task success under the operating constraints on the test data. Although most methods achieve task success rates near 97\%, their feasibility-weighted success differs substantially. TRPO-Lag achieves approximately 90\%, compared with 85\% for PPO-Lag, 73\% for CPO, and 64\% for TRAH, corresponding to relative improvements of approximately 6\%, 23\%, and 41\%, respectively. This favorable balance between task success and feasibility is consistent with the aggregate benchmark findings in~\cite{ray2019benchmarking} and can be partly attributed to the combination of KL-constrained policy updates and adaptive constraint penalties. 
Unconstrained PPO achieves only approximately 4\% feasibility-weighted success, while TRPO, TAH, and Random yield approximately zero. This gap highlights the importance of explicitly accounting for operating constraints when optimizing task success. 

\begin{figure}[h!]
    \centering
    \includegraphics[width=1\columnwidth]{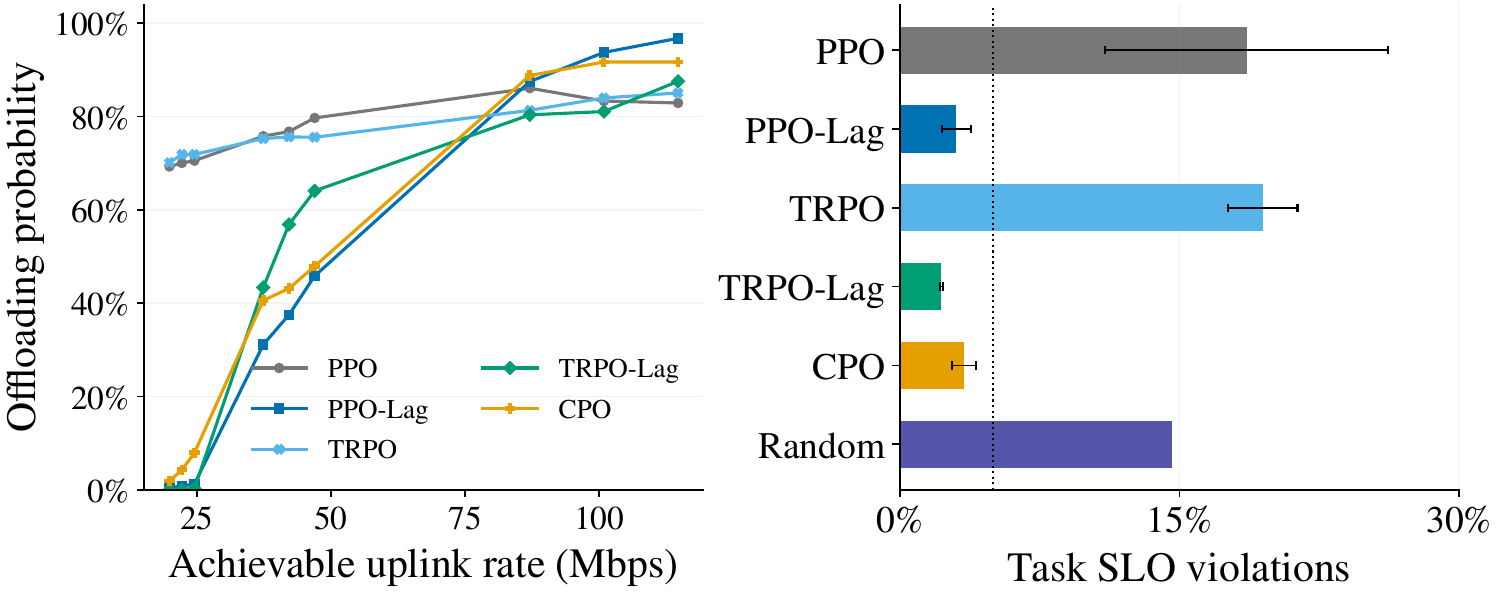}
    \caption{Offloading Behavior and SLO violations}
    \label{fig:policytaskbehaviourandrate}
    \vspace{-10pt}
\end{figure}
In terms of offloading policy, Fig. \ref{fig:policytaskbehaviourandrate} illustrates adaptation to wireless conditions. TRPO-Lag rarely offloads below approximately 25~Mbps, where communication delay limits the benefit of edge execution. Its offloading probability increases with achievable UL rate, reaching approximately 88\% at high rates, demonstrating channel-aware inference decisions. Although PPO-Lag and CPO offload more frequently in this region, TRPO-Lag achieves the lowest overall SLO violation rate, approximately 2.2\%, compared with approximately 3\% for PPO-Lag and 3.4\% for CPO. PPO and TRPO offload frequently even at low rates and incur violation rates near 19\%. 
\begin{figure}[h!]
    \centering
    \includegraphics[width=1\columnwidth]{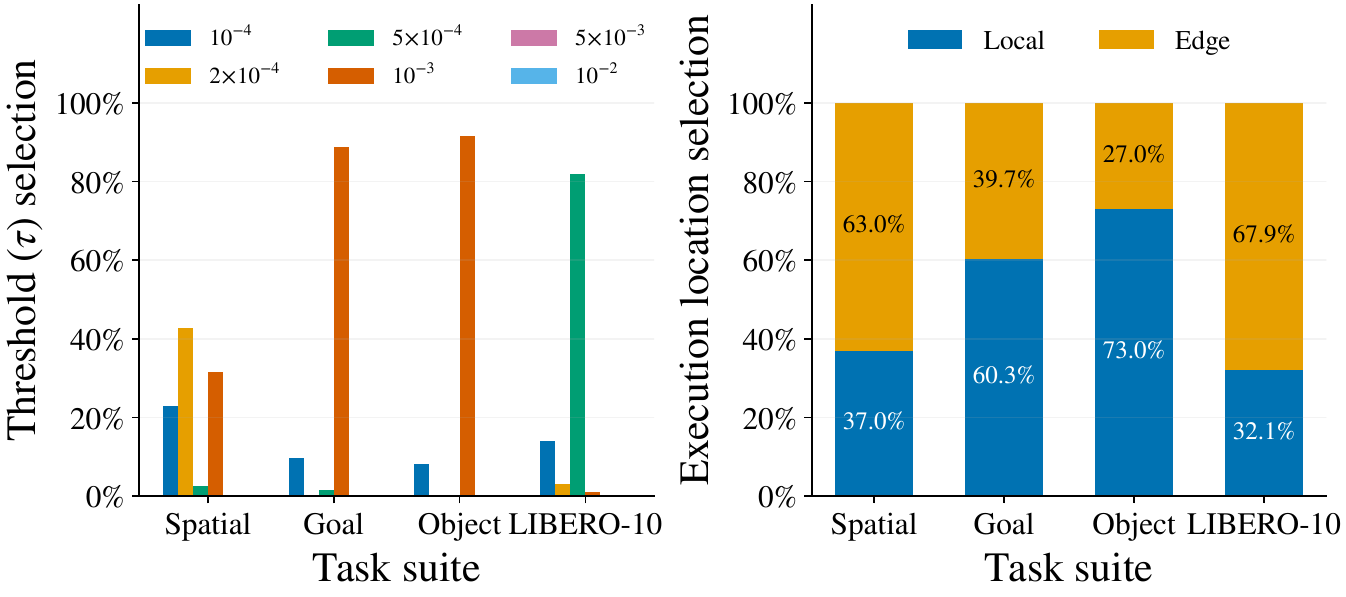}
    \caption{TRPO-Lag action distribution by task suite}
    \label{fig:policytaskbehaviour}
    \vspace{-0.4cm}
\end{figure}
Fig. \ref{fig:policytaskbehaviour} explains the learned suite-dependent decisions through the offline characteristics in Fig. \ref{fig:offlinedatasetcharacteristics}. TRPO-Lag predominantly selects $\tau=10^{-3}$ for Goal and Object, preserving high task success while requiring fewer recurrent iterations than stricter thresholds. For LIBERO-10, which comprises longer-horizon tasks, it selects $\tau=5\times10^{-4}$ in approximately 82\% of requests, consistent with the substantial success degradation observed at larger thresholds. 
Location selection complements these decisions: Spatial and LIBERO-10 are offloaded in 63.0\% and 67.9\% of requests, whereas Goal and Object are primarily executed locally. These results indicate that TRPO-Lag also learns suite-dependent operating points that preserve task success while reducing inference latency and managing shared edge resources.

\vspace{-0.3cm}
\section{CONCLUSION}
In conclusion, this paper proposed the WEALR framework for adaptive latent reasoning of VLA models in wireless edge-assisted robotic networks. By formulating the joint optimization of execution location and latent convergence threshold as a CMDP, our Lagrangian-CDRL approach dynamically balances task success against PAD energy budgets, shared edge-reasoning limits, and latency SLO constraints under time-varying wireless conditions. Evaluations across LIBERO manipulation tasks with profiled hardware costs and realistic channel models demonstrate that the proposed framework effectively satisfies resource constraints while achieving up to a 23\% gain in feasibility-weighted task success rate over baselines. Future work may explore concurrent robotic task requests with joint wireless and reasoning optimization.
\vspace{-5pt}

\section*{ACKNOWLEDGEMENT}
This work has been supported by the NSERC Canada Research Chairs Program. 

\vspace{-5pt}
\renewcommand\refname{References}
\addcontentsline{toc}{section}{References}
\bibliographystyle{IEEEtran}
\bibliography{ref}

\end{document}